\documentclass[11pt,a4paper]{article}
\usepackage[hyperref]{eacl2021}
\usepackage{times}
\usepackage{latexsym}

\usepackage{url}
\usepackage{latexsym}
\usepackage{bm}
\usepackage{amssymb}
\usepackage{array}
\usepackage{amsmath}
\usepackage{graphicx}
\usepackage{subcaption}
\usepackage{caption}

\aclfinalcopy 

\newcommand{\x}{\bm{x}}
\newcommand{\z}{\bm{z}}

\usepackage{microtype}

\title{Polarized-VAE: Proximity Based Disentangled Representation Learning for Text Generation}

\author{Vikash Balasubramanian\textsuperscript{1}\footnotemark[1], Ivan Kobyzev\textsuperscript{2}, Hareesh Bahuleyan\textsuperscript{2}, Ilya Shapiro\textsuperscript{3}, Olga Vechtomova\textsuperscript{1} \\
{\textsuperscript{1}University of Waterloo, \textsuperscript{2}Borealis AI, \textsuperscript{2}University of Windsor} \\
\tt{\{v9balasu,ovechtom\}@uwaterloo.ca} \\
\tt{ivan.kobyzev@borealisai.com} \\
\tt{hareeshbahuleyan@gmail.com} \\
\tt{ishapiro@uwindsor.ca} \\
}

\date{}

\begin{document}
\maketitle
\begin{abstract}
Learning disentangled representations of real-world data is a challenging open problem. Most previous methods have focused on either supervised approaches which use attribute labels or unsupervised approaches that manipulate the factorization in the latent space of models such as the variational autoencoder (VAE) by training with task-specific losses. In this work, we propose polarized-VAE, an approach that disentangles select attributes in the latent space based on proximity measures reflecting the similarity between data points with respect to these attributes.
We apply our method to disentangle the semantics and syntax of sentences and carry out transfer experiments. Polarized-VAE outperforms the VAE baseline and is competitive with state-of-the-art approaches, while being more a general framework that is applicable to other attribute disentanglement tasks.
\end{abstract}

\section{Introduction}
Learning representations of real-world data using deep neural networks has accelerated research within a number of fields including computer vision 
and natural language processing \cite{zhang2018survey}. Previous work has advocated for the importance of learning \emph{disentangled representations} \cite{bengio2013representation,tschannen2018recent}. Although attempts have been made to formally define disentangled representations \cite{higgins2018definition}, there is no widely accepted definition of disentanglement. However, the general consensus is that a disentangled representation should separate the distinct factors of variation that explain the data \cite{bengio2013representation}.
Intuitively, a greater level of interpretability can be achieved when independent latent units are used to encode different 
attributes of the data \cite{burgess2018understanding}. 

However, recovering and separating all the distinct factors of variation in the data is a challenging problem. For real-world datasets, there may not be a way to separate each factor of variation into a single dimension in the learnt fixed-size vector representation. An easier problem would be to separate complex factors of interest into distinct subspaces of the learnt representation. For instance, a representation for text could be separated into \textit{content} and \textit{style} subspaces, which then enables style transfer. 

Unsupervised disentanglement of underlying factors using variational autoencoders \cite{kingma2013auto} has been explored in previous work \cite{Higgins2017betaVAELB,kim2018tc}. However, \newcite{locatello2018challenging} argue that completely unsupervised disentanglement of the underlying factors may be impossible without supervision or inductive biases. 
Disentangling textual attributes in a completely unsupervised manner has been shown to be especially difficult,
but attempts have been made to leverage it for controllable text generation \cite{xu2019variational}.


In this work, we 
propose an approach referred to as polarized-VAE\footnote{The code is available at \url{https://github.com/vikigenius/prox_vae}} to disentangle the latent space into subspaces corresponding to different factors of variation. We control the relative location of representations in a particular latent subspace based on the similarity of their respective input data points according to a defined criterion (that corresponds to an attribute in the input space, e.g., syntax). This encourages similar points to be grouped together and dissimilar points to be farther away from each other in that subspace. Figuratively, we \textit{polarize} the latent subspaces, and hence the name.  

Most previous work on supervised disentanglement for text has focused on adversarial training 
\cite{john2019disentangled,yang2018stdisc}. 
Recently, the task of disentangling textual semantics and syntax into distinct subspaces has received attention from researchers.  For instance, \newcite{chen2019multi} use a sentence VAE model with several multitask losses such as paraphrase loss and word position loss for this disentanglement task.
\newcite{bao2019generating} incorporate adversarial training and make use of syntax trees along with specific multitask losses to disentangle semantics and syntax. 

In polarized-VAE, we achieve disentanglement through distance based learning. In contrast to previous approaches, our method does not require the use of several multitask losses or adversarial training, both of which can result in optimization challenges. Furthermore, we do not need precise attribute labels, and we show that using proxy labels based on the concept of similarity is sufficient. 

In summary, the main contributions of this paper are three-fold: (1) We propose a general framework for learning disentangled representations. Even though we test our method on an NLP task, the underlying concept is very general and can be applied to other domains such as computer vision; (2) We provide a method for disentanglement that does not rely on adversarial training or specialized multitask losses;
(3) We demonstrate an application of our method by disentangling the latent space into subspaces corresponding to syntax and semantics. Such a setting can be used to perform controlled text decoding such as generating a paraphrase with a desired sentence structure.

\section{Proposed Approach}
In VAEs, a probabilistic encoder $q_\phi(\z|\x)$ is used to encode  a sentence $\x$ into a   latent variable $\z$,  and a probabilistic decoder $p_\theta(\x|\z)$ attempts to reconstruct the original sentence  $\x$ from its latent representation $\z$. The objective is to minimize the following loss function:
\vspace{-0.6em}
\begin{align}
 \mathcal{L}_{\mathrm{vae}} = \mathcal{L}_{\mathrm{rec}} + \lambda_{\mathrm{kl}}\mathcal{L}_{\mathrm{kl}}   
\end{align}
where $\mathcal{L}_{\mathrm{rec}} = -\mathbb{E}_{q_\phi(\z|\x)}[\log p_\theta(\x|\z)]$~is~the sentence reconstruction loss and  $\mathcal{L}_{\mathrm{kl}} = \mathcal{D}_{\mathrm{kl}}(q_\phi(\z|\x)||p(\z))$ is the Kullback-Leibler (KL) divergence loss. The KL term ensures that the approximate posterior $q_\phi(\z|\x)$ is close to the prior $p(\z)$, which is typically assumed to be the standard normal $\mathcal{N}(\mathbf{0}, \mathbf{I})$; $\lambda_{\mathrm{kl}}$ is a hyperparameter that controls the extent of KL regularization.



\begin{figure*}
\centering
\begin{minipage}{.6\textwidth}
  \centering
  \includegraphics[width=0.9\linewidth]{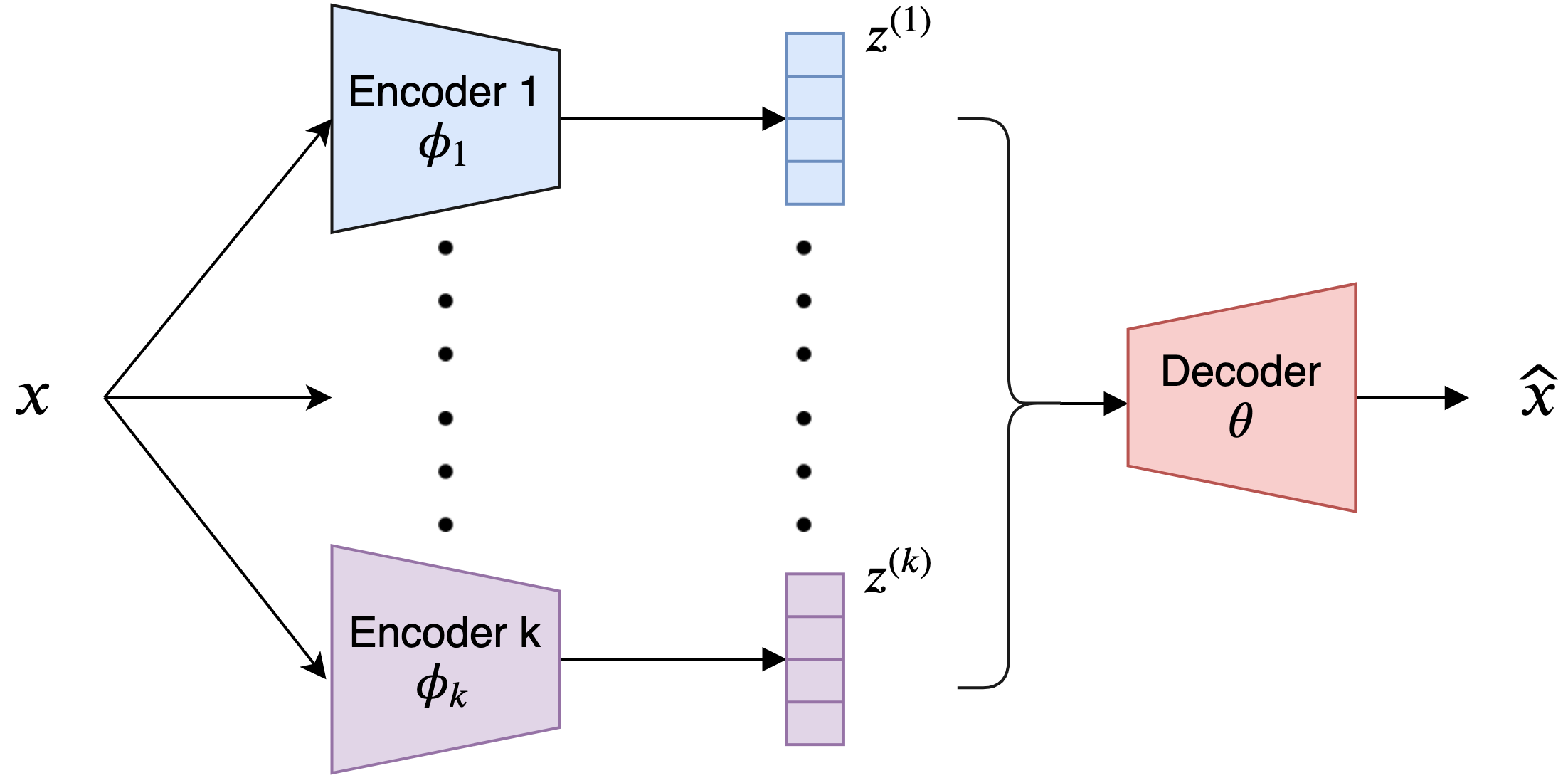}
  \captionof{figure}{Model Architecture}
  \label{fig:model}
\end{minipage}%
\begin{minipage}{.4\textwidth}
  \centering
  \includegraphics[width=0.965\linewidth]{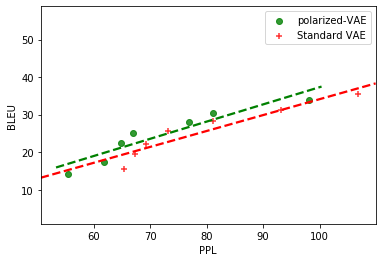}
  \captionof{figure}{BLEU vs. PPL trade-off}
  \label{fig:pplbleu}
\end{minipage}
\end{figure*}

The idea behind our polarized-VAE approach is to impose additional proximity regularization on the latent subspaces learnt by VAEs.
Let $C = \{c_1, ..., c_k\}$ be the collection of criteria, based on which we wish to disentangle the latent space $\z$ of the VAE into $k$ subspaces: $\z = [\z^{(1)}, \dots, \z^{(k)}]$. Here $\z^{(i)}$ denotes the latent subspace corresponding to the criterion $c_i$ (see Figure~\ref{fig:model}). 
In this paper, we focus on the case where the latent space is disentangled into  semantics ($c_1$) and syntax ($c_2$), i.e., $k=2$. 

\subsection{Supervision based on Similarity}
We assume that we have information (possibly noisy) about pairwise similarities of the input sentences. Given a pair of sentences, the similarity information can be either a binary label (whether both the sentences belong to the same class or not) or an integer or continuous scalar variable (e.g., edit distance). In this work, the similarity criterion is  a binary label:
\vspace{-0.6em}
\begin{equation}
    \text{Sim}(\x_i, \x_j | c)=
    \begin{cases}
      1, & \text{if } \x_i \text{ and } \x_j \text{ are similar}\\ & \text{ w.r.t. the criterion } c \in C\\
      0, & \text{otherwise}
    \end{cases}
\end{equation}

\noindent In our case, the two criteria for disentanglement are semantics ($c_1$) and syntax ($c_2$). We use this additional information to regularize the latent space of the VAE by incorporating the proximity based loss functions, denoted as  $D(\z_i^{(1)}, \z_j^{(1)} | c_1)$ and $D(\z_i^{(2)}, \z_j^{(2)} | c_2)$.

\subsection{Training Method and Proximity Function}
Extending the traditional VAE approach, we have a set of RNN-based encoders parameterized by $\phi_c$ that learn the approximate posteriors $q_{\phi_c}(\z^{(c)}|\x)$.
Given two data points $\x_i$ and $\x_j$, we denote the proximity of their encodings in the latent subspace by $D(q_{\phi_c}(\z^{(c)}|\x_i), q_{\phi_c}(\z^{(c)}|\x_j))$.
We experiment with multiple forms of proximity functions and found cosine distance to perform the best:
\vspace{-0.7em}
\begin{align}
    D(q_{\phi_c}(\z|\x_i), q_{\phi_c}(\z|\x_j)) &= d_c(\z_i, \z_j)\\
    &= \frac{1}{2}\left(1 - \frac{\z_i \z_j}{||\z_i||||\z_j||}\right) \nonumber
\end{align}

\noindent Based on the above distance, we add a regularization term to the VAE loss function as follows. For each example $(\x, c)$, we have a positive sample $\x_p$ and $m$ negative samples $\x_{n_1}, ..., 
\x_{n_m}$, such that $\text{Sim}(\x, \x_p | c) = 1$ and $\text{Sim}(\x, \x_{n_j} | c) = 0;\, j\in\{1,...,m\}$:
\vspace{-1.0em}
\begin{align}
 \mathcal{L}_{c} = \max(0, 1 + d_c(\z, \z_p) - \frac{1}{m} \sum_{j=1}^m d_c(\z, \z_{n_j})  )
\end{align}
This regularization function can be viewed as a max-margin loss over the proximity function.
The final objective then becomes
\vspace{-1.0em}
\begin{equation}
    \mathcal{L} = \mathcal{L}_{\mathrm{vae}} + \sum_{c \in C} \lambda_{c}\mathcal{L}_c
\end{equation}

\section{Experiments}
To demonstrate the effectiveness of polarized-VAE in obtaining disentangled representations, we carry out semantics-syntax separation of textual data, using the Stanford Natural Language Inference  dataset (SNLI,  \newcite{bowman2015large}). Model implementation details are provided in Appendix ~\ref{sec:imp-details}.

\subsection{Reconstruction and Sample Quality}
We evaluate our model on reconstruction and sample quality to ensure that the distance-based regularization used does not adversely impact its reconstruction or sampling capabilities. For this purpose, we  compare our model and the standard VAE on two metrics: reconstruction BLEU \cite{Papineni2002bleu} and the Forward Perplexity (PPL)\footnote{PPL is computed using the KenLM toolkit \cite{kenlm}} \cite{zhao2018adversarially} of the generated sentences obtained by sampling from the model's latent space.  As seen in Figure \ref{fig:pplbleu}, there is a clear trade-off between reconstruction quality and sample quality, which is expected. Overall, polarized-VAE performs slightly better than standard VAE and this indicates that the proximity-based regularization does not inhibit the model capabilities.


\subsection{Controlled Generation and Transfer}
We follow previous work  \cite{mchen-controllable-19,bao2019generating} and
analyze the performance of controlled generation by evaluating syntax transfer in generated text. 
Given two sentences, $\x_{\mathrm{sem}}$ and $\x_{\mathrm{syn}}$, 
we wish to generate a third sentence that combines the semantics of $\x_{\mathrm{sem}}$ and the syntax of $\x_{\mathrm{syn}}$ using the following procedure:
\begin{align*}
    \z_{\mathrm{sem}} \sim q_{\phi_1}(\z^{(1)} | \x_{\mathrm{sem}})\hspace{0.5em}&; \hspace{0.5em}
    \z_{\mathrm{syn}} \sim q_{\phi_2}(\z^{(2)} | \x_{\mathrm{syn}}) \\
    \z = [\z_{\mathrm{sem}}, \z_{\mathrm{syn}}] \hspace{0.5em}&; \hspace{0.5em}
    \x \sim p_\theta(\x | \z) 
\end{align*}

\noindent Following the evaluation methodology of \newcite{bao2019generating}, we measure transfer based on (1) semantic content preservation for the semantic subspace  and (2) the tree edit distance \cite{zhang} for the syntactic subspace.

We consider pairs of sentences from the SNLI test set for evaluation. We would like the generated sentence to be close to $\x_{\mathrm{sem}}$ and different from $\x_{\mathrm{syn}}$ in terms of semantics, which is measured using BLEU scores. We also report the difference to indicate the strength of transfer denoted by $\Delta$BLEU. Additionally, we would like the generated sentence to be syntactically similar to 
$\x_{\mathrm{syn}}$ and different from $\x_{\mathrm{sem}}$, which is measured by averaged sentence-level Tree Edit Distance (TED). We also report $\Delta$TED to indicate the strength of the syntax transfer. Finally, we use the Geometric Mean of $\Delta$BLEU and $\Delta$TED to report a combined score $\Delta$GM. 

\begin{table*}[!t]
\centering
\small
\setlength\tabcolsep{3pt}
\begin{tabular}{|c | c  c | c | c c | c | c | c c c|} 
 \hline
 & \multicolumn{2}{c|}{\textbf{BLEU}} & & \multicolumn{2}{c|}{\textbf{TED}} & & & \multicolumn{3}{c|}{\textbf{Human Eval $(\%)$} }\\ 
 Model & $\x_{\mathrm{sem}}$\textsuperscript{$\uparrow$} & $\x_{\mathrm{syn}}$\textsuperscript{$\downarrow$}& $\Delta$BLEU\textsuperscript{$\uparrow$} & $\x_{\mathrm{sem}}$\textsuperscript{$\uparrow$} & $\x_{\mathrm{syn}}$\textsuperscript{$\downarrow$} & $\Delta$TED\textsuperscript{$\uparrow$}& $\Delta$GM\textsuperscript{$\uparrow$} & $\mathrm{sem}$ & $\mathrm{syn}$ & $\mathrm{fluency}$ \\ [0.5ex] 
 \hline\hline
 Standard VAE & 4.75 & 4.67 & 0.08 & 13.70 & 13.60 & 0.10 & 0.28 & 11 & 11 & \textbf{43}\\
 \newcite{bao2019generating} & \textbf{13.74} & 6.15 & 7.59 & \textbf{16.19} & 13.10 & \textbf{3.08} & 4.83 & 24 & \textbf{58} & 19\\
 \hline
 polarized-VAE & 10.78 & 0.92 & \textbf{9.86} & 14.09 & 11.67 & 2.42 & \textbf{4.88} & \textbf{65} & 31 & 38\\
 polarized-VAE (\texttt{wo}) & 9.82 & 0.84 & 8.98 & 14.12 & 11.65 & 2.47 & 4.71 & - & - & -\\
 polarized-VAE (\texttt{len}) & 10.10 & \textbf{0.76} & 9.34 & 12.68 & \textbf{11.44} & 1.44 & 3.67 & - & - & -\\
 polarized-VAE (\texttt{wo}, \texttt{len}) & 9.41 & 0.87 & 8.54 & 12.65 & 11.48 & 1.17 & 3.16 & - & - & -\\
 \hline
\end{tabular}
\vspace{-0.5em}
\caption{Syntax transfer results on SNLI. \newcite{bao2019generating} report TED after multiplying by 10, we report their score after correction. For each model, the human evaluation scores represent percentage of instances that it was ranked the best for a given criteria (semantics preservation/syntax transfer/fluency).}
\vspace{-1.0em}
\label{table:1}
\end{table*}

Our default variant of polarized-VAE uses the entailment labels from SNLI dataset as a proxy for semantic similarity based on which positive and negative samples are chosen. For this model, we threshold the difference in TED of syntax parses as a proxy for syntactic similarity. 
As shown in Table~\ref{table:1}, we also evaluate three other variants of our model. In polarized-VAE (\texttt{wo}) we use word overlap (BLEU scores) as a heuristic proxy for estimating semantic similarity, while keeping syntactic training unchanged. We also experiment with heuristics for syntax in polarized-VAE (\texttt{len}) where we use length as a heuristic proxy for syntax, while still using ground truth entailment labels for semantic training. Finally we combine these two heuristics in polarized-VAE (\texttt{wo}, \texttt{len}), which can be viewed as an unsupervised variant that does not make use of any ground truth labels or syntax trees.

Our model outperforms the VAE baseline on all metrics. In comparison to \cite{bao2019generating}, polarized-VAE is much better at ignoring the semantic information present in $\x_{\mathrm{syn}}$ during syntax transfer, as evidenced by our lower BLEU scores w.r.t. $\x_{\mathrm{syn}}$. On the other hand, we perform slightly worse on BLEU w.r.t. $\x_{\mathrm{sem}}$. Our model does a better job at matching the syntax of sentence $\x_{\mathrm{syn}}$ as indicated by the lower TED score w.r.t. $\x_{\mathrm{syn}}$. Qualitative samples of syntax transfer are provided in Appendix
~\ref{sec:transfer-examples}.

\subsection{Human Evaluation}
\label{ssec:human-eval}
We carried out a human evaluation study for comparing outputs generated from different models. The test setup is as follows - we provide as input two sentences, $x_{\mathrm{sem}}$ and $x_{\mathrm{syn}}$ to the model; we wish to generate a sentence that combines the semantics of $x_{\mathrm{sem}}$ and the syntax of $x_{\mathrm{syn}}$. 
We asked 5 human annotators to evaluate the outputs from the 3 models: baseline-VAE, polarized-VAE and the model from \cite{bao2019generating}.

Each annotator was shown the input sentences ($x_{\mathrm{sem}}$ and $x_{\mathrm{syn}}$) and the outputs from the 3 models (randomized so that the evaluator is unaware of which output corresponds to which model). They were then asked to pick the one best output for each of the following three criteria: (1) semantic preservation --- level of semantic similarity with respect to $x_{\mathrm{sem}}$, (2) syntactic transfer  --- level of syntactic similarity with respect to $x_{\mathrm{syn}}$ and (3) fluency. We obtained annotations on 100 test set examples from SNLI dataset.
To aggregate the annotations, we used majority voting with manual tie breaking to find the best model for each test example (and for each test criteria). 


For each model, we report the percentage of instances where it was voted as best for each criteria. From the human evaluation results in Table~\ref{table:1}, we note that polarized-VAE is better at semantic transfer and worse at syntactic transfer in comparison to \cite{bao2019generating}. The human evaluation results are consistent with the automatic evaluation metrics, where polarized-VAE scores higher on $\Delta$BLEU (indicator of semantic transfer strength) and \cite{bao2019generating} is better at $\Delta$TED (indicator of syntax transfer strength).
With respect to fluency criterion, polarized-VAE ranks higher than \cite{bao2019generating}. However, the most fluent sentences are produced by the baseline VAE. We hypothesise this to be due to the presence of additional regularization terms in the loss functions of both \cite{bao2019generating} and polarized-VAE, which in turn affects the fluency of their generated text (due to the deviation from the reconstruction objective).

\section{Conclusion and Future Work}
We proposed a general approach for disentangling latent representations into subspaces using proximity functions. Given a pair of data points, a predefined similarity criterion in the original input space determines their relative distance in the corresponding latent subspace, which is modelled via a proximity function. 
We apply our approach to the task of disentangling semantics and syntax in text. Our model substantially outperforms the VAE baseline and is competitive with the state-of-the-art approach while being more general as we do not use specific multitask losses or architectures to encourage preservation of semantic or syntactic information. Our methodology is orthogonal to the multitask learning approaches by \newcite{chen2019multi} and \newcite{bao2019generating} and can be naturally combined with their methods. We would further like to investigate this approach on disentanglement applications outside of NLP. Another interesting research direction would be to further explore suitable proximity functions and identify their properties that could facilitate disentanglement.

\section{Acknowledgements}
We would like to thank Dr. Pascal Poupart for his valuable insights and ideas. We would also like to thank Compute Canada (\url{www.computecanada.ca}) for their support and GPU resourcees.

This Research was funded by the MITACS Accelerate program in collaboration with Borealis AI.

\bibliography{anthology,eacl2021}
\bibliographystyle{acl_natbib}

\clearpage
\appendix

\section*{\centering{\Large{Appendix}}}
\section*{\centering{\Large{Polarized-VAE: Proximity Based Disentangled Representation Learning for Text Generation}}}

\section{Model and Training Details}
\label{sec:imp-details}
Both the semantic and syntactic encoders are bidirectional LSTMs \cite{Hochreiter1997} with hidden size of 128, followed by two feed-forward layers to parameterize the Gaussian mean ($\mu$) and standard deviation ($\sigma$) parameters similar to standard VAE formulations used by \cite{bao2019generating}. The latent space dimensions were taken to be $dim(z^{1}) = 64$ and $dim(z^{2}) = 16$. The decoder is a unidirectional LSTM with a hidden size of 128. We train the model for 30 epochs in total using the ADAM optimizer \cite{kingma2014adam} with the default parameters and a learning rate of $0.001$.

We adopt the standard tricks for VAE training including dropout and KL annealing followed by \cite{Bowman_2016}. We anneal both semantic and syntactic KL weights ($\lambda_{kl}$) upto 0.3 (5000 steps) using the same \texttt{sigmoid} schedule \cite{nlgnvm}.

\section{Proximity Functions}
We provide results for the other proximity functions that we explored for the polarized-VAE model.

\begin{table}[htb]
\small
\centering
\begin{tabular}{| c | c | c | c |} 
 \hline
Metric &
$\Delta$BLEU\textsuperscript{$\uparrow$} & $\Delta$TED\textsuperscript{$\uparrow$}& $\Delta$GM\textsuperscript{$\uparrow$}\\ [0.5ex] 
 \hline
 Cosine Distance & \textbf{9.86} & \textbf{2.42} & \textbf{4.88}\\
 Hellinger Distance & 4.12 & 0.86 & 1.42\\
 MMD & 5.21 & 1.17 & 1.91\\
 KL Divergence & 4.32 & 0.75 & 1.28\\
 JS Divergence & 5.81 & 1.46 & 2.33\\
 \hline
\end{tabular}
\caption{Comparison of polarized-VAE with different proximity functions.}
\label{table:prox}
\end{table}

\noindent We note that since there is no closed form expression for the JS divergence between two Normal Random variables we used the generalized JS Divergence proposed by \cite{nielsen2019generalization}.

\section{Transfer Examples}
\label{sec:transfer-examples}
We provide qualitative examples of our transfer experiments, where we generate a sentence with the semantics of $x_{\mathrm{sem}}$ and the syntactic structure of $x_{\mathrm{syn}}$ in Table \ref{tab:transfer_exmaples}. We also provide the sentences generated by a standard-VAE for comparison.

\section{Disentanglement of Latent Subspaces}
We test if there a possibility that the two latent subspaces encode similar information. This is only likely to happen if the attributes themselves are highly correlated (e.g., if we want to disentangle syntax from length). For such cases, even existing methods based on adversarial disentanglement \cite{john2019disentangled} may fail to completely separate out correlated information.

However, if the attributes are different enough (or ideally independent) for e.g., syntax and semantics, this is less problematic. Note that we apply our proximity loss independently to each of the sub-spaces (i.e., leaving the other space(s) untouched for a given input). This encourages the semantic encoder to encode semantically similar sentences close together and dissimilar ones far apart in the semantic space (same applies for the syntax encoder).

We empirically compute correlations between the semantic and syntax latent vectors for 1000 test sentences as a way to check whether the two encoders learn similar information. By feeding 1000 sentences from the test set to the Polarized-VAE, we obtain their corresponding semantic ($z_{\mathrm{sem}}$) and syntax ($z_{\mathrm{syn}}$) latent vectors. 
We then empirically compute the correlation between $z_{\mathrm{sem}}$ and $z_{\mathrm{syn}}$. To analyze the level of similarity of information represented in $z_{\mathrm{sem}}$ and $z_{\mathrm{syn}}$, we report the maximum absolute correlation (max across all pairs of dimensions) and also the mean absolute correlation. A higher value of correlation would indicate that there is more overlapping information learnt by the semantic and syntactic encoders. 
As illustrated in Table~\ref{table:disentanglement}, the analysis indicates that the semantic and syntax latent vectors in polarized-VAE encodes less correlated information than standard-VAE (due to the proximity-based regularization). This demonstrates that the 2 latent spaces learned by our model encode sufficiently different information.

\begin{table}[!ht]
\small
\centering
\begin{tabular}{|c|c|c|}
\hline
Model & Max Abs Corr\textsuperscript{$\downarrow$} & Mean Abs Corr\textsuperscript{$\downarrow$}\\
\hline
standard-VAE & 0.62 & 0.1\\
polarized-VAE & \textbf{0.25} & \textbf{0.05}\\
\hline
\end{tabular}

\caption{Maximum Absolute Correlation and Mean Absolute Correlation between the semantic and syntactic latent vectors.}
\label{table:disentanglement}
\end{table}

\begin{table*}[!ht]
    \centering
    \begin{tabular}{p{0.23\linewidth}|p{0.22\linewidth}|p{0.22\linewidth}|p{0.22\linewidth}}
        \hline
        \hspace{3em} \bm{$x_{\mathrm{sem}}$} & 
        \hspace{3em} \bm{$x_{\mathrm{syn}}$} & 
        \textbf{\hspace{1em} polarized-VAE} & \textbf{\hspace{1em} standard-VAE} \\
        \hline
        A man works near a vehicle. & A woman showing her face from something to her friend. & A man directing traffic on a bicycle to an emergency vehicle. & A woman works on a loom while sitting outside.\\
        \hline
        A family in a party preparing food and enjoying a meal. & Man reading a book. & A person enjoying food. & A man plays his guitar.\\
        \hline
        Two young boys are standing around a camera outdoors. & Three kids are on stage with a vacuum cleaner. & Two young boys are standing around a camera outdoors. & Two people are standing on a snowy hill.\\        
        \hline
        There are a group of people sitting down. & They are outside. & There are people. & They are outside\\
        \hline
        a woman wearing a hat and hat is chopping coconuts with machete. & The person is in a blue shirt playing with a ball. & a woman with a hat is hanging upside down over utensils. & A girl in a pink shirt and elbow pads is swirling bubbles.\\
        \hline
        The young girl and a grownup are standing around a table , in front of a fence. & A guy stands with cane outdoors. & The young girl is outside. & The little boy is doing a show.\\
        \hline
        A person is sleeping on bed. & A man and his son are walking to the beach , looking for something. & A man and a child sit on the ground covered in bed with rocks. & A man is wearing blue jeans and a blue shirt walking.\\
        \hline
        The men and women are enjoying a waterfall. & A dog is holding an object. & The man and woman are outdoors. & The two men are working on the roof.\\
        \hline
        a man dressed in uniform. & There is a man with a horse on it. & A man dressed in black clothing works in a house. & A man dressed in black and white holding a baby.\\
        \hline
    \end{tabular}
    \caption{Examples of transferred sentences that use the semantics of $x_{sem}$ and syntax of $x_{syn}$}
    \label{tab:transfer_exmaples}
\end{table*}
\end{document}